\providecommand{\reserveinserts}[1]{}
\documentclass[HARVARD,LATO2COL]{WileyNJDv5}
\usepackage{amsmath}
\usepackage{mathtools}
\usepackage{amsfonts}
\usepackage{enumitem}
\usepackage{threeparttable}
\usepackage{subcaption}
\usepackage{tabularx}  
\usepackage{booktabs}  
\usepackage{multirow}  
\usepackage{makecell}  
\usepackage{siunitx}
\usepackage{ragged2e}  
\usepackage{textcomp}
\usepackage{hyperref}
\articletype{Article Type}%
\received{Date Month Year}
\revised{Date Month Year}
\accepted{Date Month Year}
\journal{Computer-Aided Civil and Infrastructure Engineering}
\volume{00}
\copyyear{2025}
\startpage{1}

\newcommand{\mytitle}{3D Modeling and Automated Measurement of Concrete Cracks via Segment Anything Refinement and Visual Inertial LiDAR Fusion}  

\begin{document}

\title{\mytitle}

\author[1,2]{Pengru Deng*}

\author[1]{Jiapeng Yao}

\author[3]{Chun Li}

\author[4]{Su Wang}

\author[5]{Xinrun Li*}

\author[5]{Varun Ojha}

\author[1,2]{Xuhui He}

\authormark{Deng \textsc{et al.}}
\titlemark{\mytitle}

\address[1]{\orgdiv{School of Civil Engineering}, \orgname{Central South University}, \orgaddress{\state{Changsha}, \country{China}}}

\address[2]{\orgdiv{Hunan Provincial Key Laboratory for Disaster Prevention and Mitigation of Rail Transit Engineering Structures},{\state{Changsha}, \country{China}}}

\address[3]{\orgname{Carizon}, \orgaddress{\state{Shanghai}, \country{China}}}

\address[4]{\orgname{Nvidia}, \orgaddress{\state{Shanghai}, \country{China}}}

\address[5]{\orgdiv{School of Computing}, \orgname{Newcastle University}, \orgaddress{\state{Tyne and Wear}, \country{United Kingdom}}}

\corres{Pengru Deng, Central South University, No. 932 South Lushan Road, Changsha Hunan 410083, P.R. China.
\email{pengrudeng@csu.edu.cn}\\
\\
Xinrun Li, School of Computing, Newcastle University, Tyne and Wear, NE4 5TG, United Kingdom.
\email{xinrun.li@ncl.ac.uk}}

\abstract[Abstract]{In practical applications, AI-based concrete crack inspection still suffers from performance degradation in few-shot, unfamiliar scenarios and lacks the capability for high-precision, synchronized quantification of three-dimensional (3D) crack geometry and location without manual post-processing.
To address these limitations, a systematic methodology for crack segmentation, 3D reconstruction, and automated measurement is proposed, grounded in computer vision and Simultaneous Localization and Mapping (SLAM) techniques.
First, a novel prompt generation strategy and a tailored segmentation quality assessment module are developed to improve the performance of the Segment Anything Model (SAM), enabling few-shot crack segmentation with strong generalization across diverse and unseen scenarios.
Second, a comprehensive concrete cracks reconstruction within a 3D representation is achieved through a newly proposed Visual Inertial LiDAR (VIL) SLAM-based fusion approach. By integrating multi-frame RGB images, LiDAR point clouds, and inertial measurements, the method enables precise alignment of crack segmentation masks with 3D structural geometry, generating high-precise, dense, and semantically enriched point clouds that capture fine-grained crack details at real-world scale.
Furthermore, an automated measurement module is introduced to directly quantify detailed crack geometrical and spatial information from the established 3D representation, eliminating manual post-processing and advancing beyond traditional image-based methods. Finally, extensive experiments are successfully conducted on diverse concrete structures validating the accuracy, robustness, and effectiveness in complex, non-planar, and cluttered environments of the proposed method. The source code is available at \href{https://github.com/XR-Lee/CrackSeg}{https://github.com/XR-Lee/CrackSeg}. }


\maketitle

\section{Introduction}\label{sec1}
With the extended service life of infrastructure such as bridges and tunnels, structural aging caused by accumulated loads and environmental exposure has become a critical concern.
This increases the demand for regular inspection and maintenance \citep{SPENCER2019199}.
Among various damage symptoms, surface cracking is one of the most common and visually apparent phenomena observed during inspection and is often used as an important indicator for condition assessment \citep{yang_automatic_2018}.
Mechanically, the width and number of cracks are critical markers for assessing the structural integrity and predicting changes in permeability as the structure deteriorates. 
The spatial distribution and orientation of cracks provide insights into their types, whether they are structural or non-structural, and their underlying causes \citep{zeng_systematic_2024}. The coupled characterization of the width and position of the crack serves as a vital input for the simulation of the damage state, the stress redistribution model, and the precise evaluation of structural health. Thus, the objective and simultaneous measurement of three-dimensional (3D) geometry and distribution of cracks is a prerequisite for comprehensive crack analysis.

Manual crack inspections, particularly in elevated or complex structural components, are often inconsistent, subjective, and spatially limited \citep{SPENCER2019199}.
These limitations have driven interest in automated methods that offer scalable, consistent, and objective solutions for crack detection and measurement.
Typically, such systems perform two core tasks: 
crack identification and the quantification of geometric features such as width and position.
For these methods to be widely applicable, they must generalize well across real-world conditions and accurately integrate spatial geometry with semantic understanding of cracks.

To develop generalizable crack detection methods, research has evolved from task-specific deep learning models to foundation models. 
Inspired by the success of deep learning algorithms in medical computer vision, \cite{zhang_road_2016} pioneered the application of convolution neural networks (CNN) for crack classification in asphalt pavements. 
Since then, numerous scholars have explored the field of crack image recognition, focusing primarily on three key areas: crack detection,crack classification, and pixel-level segmentation.
Early CNN-based methods focused on basic crack classification and detection \citep{cha_deep_2017,silva_concrete_2018,chaiyasarn_crack_2018}, but they lacked precision in capturing crack morphology.
This limitation prompted a shift toward pixel-level segmentation approaches \citep{zhangCrackDetectionUsing2021}.
Subsequent work enhanced segmentation accuracy by adopting more complex neural network architectures and improved training efficiency through network structure optimization \citep{Zhouzhong2022,Zhouzhong2023,Sun2024,Chun2024}.

Despite these advances, crack segmentation is still far from mature when generalization requirements in real-world engineering scenarios are considered.
Most task-specific deep learning models, such as the crack segmentation small model, are trained on fixed datasets, and their performance often degrades markedly when applied to unseen scenarios due to data scarcity and environmental variability.
In practical engineering inspections, each task typically represents a new scene, making large-scale manual annotation and retraining impractical.
Existing methods mainly emphasize pixel-wise accuracy on benchmark datasets, while overlooking the fact that crack segmentation in real applications is inherently a cross-domain problem.
New inspection scenes usually exhibit different surface tones, textures, illumination conditions, and background interferences from non-concrete components.
Under such domain shifts, the performance of task-specific segmentation networks tends to deteriorate substantially in the absence of scenario-specific annotated data and retraining.
Therefore, in most existing practical engineering crack measurement tasks, the crack segmentation part is annotated and trained with additional image data from detection scenarios \citep{ding_crack_2023,deng_binocular_2023,zhao_intelligent_2024}.
This limitation has been widely observed in practical applications and strongly motivates the development of crack segmentation methods with improved generalization capability.

More recently, the advent of foundation models, such as Segment Anything Model (SAM), has introduced new possibilities for enhancing segmentation generalization. 
\cite{TENG2025105906} explored a fractal dimension matrix prompt with SAM to enhance segmentation generalization in local images of concrete cracks.
Although foundation models offer a new approach to this challenge, their segmentation accuracy in complex scenes is inconsistent. A key limitation is their reliance on prompts: simple image-processing-based prompts-such as those derived from a fractal dimension matrix-often lack feature specificity and activate both relevant and irrelevant features, including noise, stains, and non-concrete elements.
Therefore, achieving reliable crack segmentation in full-scene images under realistic conditions, where the surface appearance is disturbed by stains, markings, and background clutter, remains an unresolved challenge.

The next critical step in crack inspection involves the simultaneous measurement of crack width and spatial position. Accurate 3D sensing and measurement have been long recognized as an important enabler for structural assessment and monitoring \citep{park2007tls,park2015mocap,oh2017strain}. As described in Table~\ref{tab:methodology_compare}, early research efforts primarily relied on the image from the fixed camera to reconstruct crack skeletons and estimate the width \citep{shokri_semantic_2022, zhang_binocular_2023}.
With the growing availability of 3D sensing, bridge inspection research has increasingly moved toward combining 3D geometry with semantic understanding. \cite{linStructureorientedLossFunction2025} performed point cloud segmentation on the semantic information of different bridge components; \cite{linBridgeInspectionUsing2025} located corrosion damage in the reconstructed model. 
Building on this trend, crack-oriented studies have further explored 3D reconstruction techniques to enable crack localization and measurement, among which Structure-from-Motion (SfM)-based approaches have been widely investigated.
For example, \cite{kim_automated_2022} used a dual lens system that combined wide-angle and telephoto cameras to achieve the location and measurement of the crack. \citet{ding_crack_2023} utilized Unmanned Aerial Vehicles (UAVs) equipped with cameras and laser rangefinders to measure crack width, while \citet{zhao_intelligent_2024} applied UAV-based photogrammetry to generate 3D models of dam surfaces for crack data extraction.
Another prominent direction involves Simultaneous Localization and Mapping (SLAM) techniques. \citet{deng_binocular_2023} developed a binocular visual SLAM (V-SLAM) framework that enables 3D localization of cracks and measurement of their lengths.
More recently, multi-modal fusion strategies have been proposed to address the limitations of conventional methods. \cite{HU2024105262} achieved localized width estimation using single-frame fusion, relaxing the typical planarity constraints. \citet{feng_crack_2023} implemented a multi-frame multi-modal SLAM framework that achieves 3D crack localization and estimates width by extracting crack boundaries directly from high-resolution image data.

\renewcommand{\arraystretch}{1}  

\begin{table*}[!tb]
\centering
\caption{Comparative analysis of crack width measurement methods.}
\label{tab:methodology_compare}

\setlength{\tabcolsep}{4pt} 

\begin{tabularx}{\textwidth}{@{}>{\raggedright\arraybackslash}p{3.3cm}
                                 >{\raggedright\arraybackslash}p{2.5cm}
                                 >{\raggedright\arraybackslash}p{2.9cm}
                                 >{\raggedright\arraybackslash}p{2.6cm}
                                 >{\raggedright\arraybackslash}p{2.5cm}
                                 >{\raggedright\arraybackslash}p{2.2cm}@{}}
\toprule
\multirow{2}{=}{\textbf{Methods}} &
\multicolumn{2}{l}{\textbf{Crack Width Measurement Accuracy}} &
\multirow{2}{=}{\textbf{Support Non-planar Surfaces}} &
\multirow{2}{=}{\textbf{3D Position \& Width Sync}} &
\multirow{2}{=}{\textbf{Automated Measurement}} \\
\cmidrule(lr){2-3}
& \textbf{Mean error < 0.1 mm} & \textbf{Acquisition requirements} & & & \\
\midrule
\textbf{Single-frame Image} \citep{zhang_binocular_2023}  & Yes & Fixed Camera & No & No & No \\
\textbf{UAV photogrammetry} \citep{zhao_intelligent_2024} & No  & High-overlap Images & No & No & No \\
\textbf{SfM \& Reference Marker} \citep{LiuYuFei2020} & No  & Known-size Marker & No & No & No \\
\textbf{UAV \& Laser Rangefinder} \citep{ding_crack_2023}  & Yes  & Full-field Scale Calibration & No & No & No \\
\textbf{Binocular V-SLAM} \citep{deng_binocular_2023}  & No  & Only supports crack length measurement & Yes & No & No \\
\textbf{SLAM \& Single-frame Fusion} \citep{HU2024105262} & Yes  & None & Yes & No & No \\
\textbf{Multi-sensor SLAM} \citep{feng_crack_2023} & No & None & Yes & No & No \\
\textbf{Proposed Method} & \textbf{Yes}  & \textbf{None} & \textbf{Yes} & \textbf{Yes} & \textbf{Yes} \\
\bottomrule
\end{tabularx}

\vspace{0.3em}
\begin{minipage}{\textwidth}
\footnotesize
\textbf{Notes:} “Yes" indicates that the method supports the capability; “No" indicates it does not; “None" means no specific requirement for data acquisition.
The Binocular V-SLAM method supports 3D localization and crack length measurement, but does not provide width estimation.
\end{minipage}
\end{table*}

However, current methods fall short of achieving high-precision fusion between crack semantic information and 3D geometric data, thereby preventing the direct and simultaneous extraction of crack width and spatial position from a fused 3D crack semantic point cloud.
Techniques based on SfM and V-SLAM rely on bundle-adjusted feature point reconstruction but often suffer from limited geometric accuracy and substantial loss of fine image details during point cloud generation. 
Meanwhile, SfM-based methods still face significant limitations when applied to real-world engineering structures, particularly in terms of reconstruction quality, detection efficiency, and measurement constraints. 
Although SfM-based methods are applicable to general 3D reconstruction, their geometric accuracy is typically at the centimeter level, which is insufficient to meet the requirements of detecting cracks at the millimeter scale.
Furthermore, SfM, feature-point-based reconstruction methods require substantial image overlap and precise trajectory planning, posing challenges in dynamic or large-scale environments. These methods also struggle with real-time processing due to computational demands.
More importantly, crack width in SfM-based methods is usually measured in the image domain under planarity assumptions, which decouples width measurement from its true 3D spatial location \citep{ding_crack_2023,zhao_intelligent_2024}.
This limitation prevents the direct construction of damage-informed structural models that require spatially localized crack width information.
In contrast, while multi-modal SLAM methods offer improved geometric fidelity, they often fail to preserve clear semantic delineation of cracks. 
Consequently, current methods directly estimate the crack width from two-dimensional (2D) semantic cues or a single-frame point cloud, resulting in fragmented, localized measurements that cannot capture the coupled relationship between width and spatial position.

\begin{figure*}[t]
    \centering  
    \includegraphics[width=0.7\linewidth]{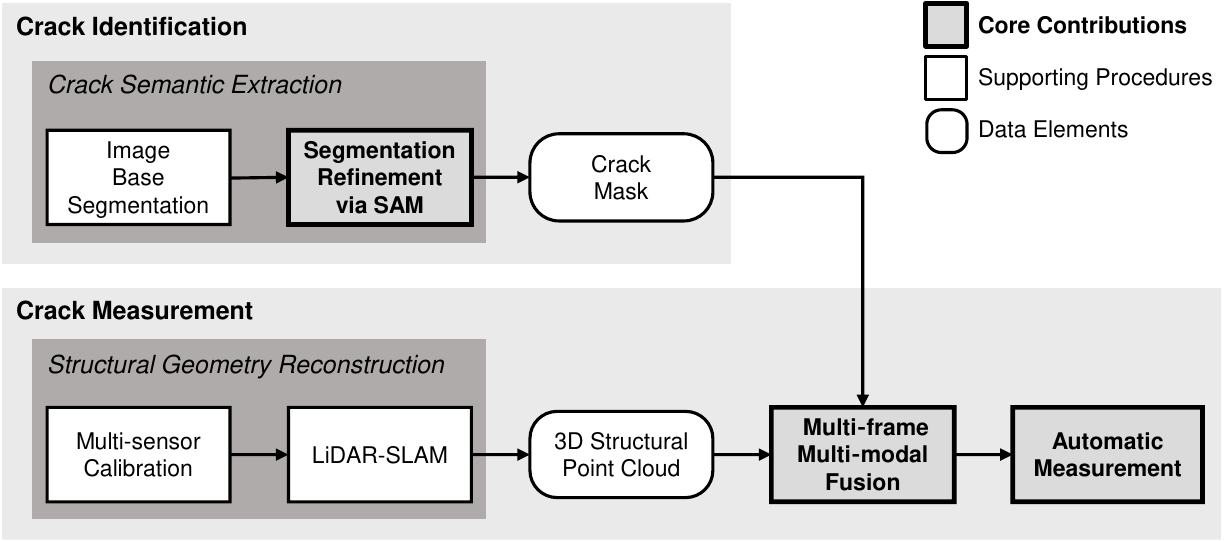}
    \caption{Overview of the proposed crack modeling \& automated measurement methodology, consisting of two key stages: identification and measurement. In the identification stage, crack masks are obtained through refinement of a foundation model SAM. These masks, along with the 3D point cloud, are integrated using a multi-frame, multi-modal fusion strategy to perform automatic measurement.}
    \label{fig:workflow-of-proposed-framework}
\end{figure*}

To address the above challenges, as shown in Figure~\ref{fig:workflow-of-proposed-framework}, this study pioneers a novel method that features high-generalization crack segmentation, 3D reconstruction of concrete with cracks, and automated extraction of 3D crack geometry and spatial information. First, the generalization capability of the segmentation foundation model SAM is effectively extended to concrete crack identification through a newly proposed prompt generation strategy that incorporates crack awareness via a dedicated deep learning model. To further enhance robustness, a segmentation quality assessment module is introduced. These improvements allow the proposed method to perform reliably even in complex crack environments. Second, a crack-oriented, multi-frame, multi-modal fusion framework is proposed based on Light Detection and Ranging (LiDAR)-SLAM reconstruction, achieving precise alignment between crack semantics and 3D geometric structures. This framework supports 3D structural reconstruction at real-world scale, enriched with high-precision semantic crack information. Furthermore, the objective and comprehensive quantification of 3D crack characteristics, such as width and spatial location, is achieved by a proposed automatic pipeline based on the 3D semantic point cloud. Finally, extensive experiments on real-world concrete structures demonstrate the effectiveness of the proposed method, yielding a 6\% improvement in Intersection over Union (IoU) compared to state-of-the-art task-specific models, and achieving a mean crack width estimation error below 0.1 mm in 3D measurement.

\section{Generalized Concrete Crack Segmentation}
Task-specific crack segmentation methods typically rely on training and testing within fixed datasets, but their performance often degrades significantly when transferred to novel scenes, necessitating additional annotations. 
In contrast, foundation models exhibit exceptional generalization abilities but heavily rely on the quality and relevance of crack prompts.
To address this gap, this study formulates a novel segmentation approach that bridges the crack-specific perception of deep learning models with the generalization capabilities of foundation models, enabling robust performance across novel scenes.
Specifically, a crack-specific segmentation prompt is developed and optimized. A sampling algorithm is designed to extract skeletal points from the base crack mask. 
These points serve as prompts for the appropriate resolution, enabling the foundation model to refine segmentation results with high precision. 
Furthermore, to address potential post-refinement issues, a quality assessment module identifies and rejects outputs that do not align with the input topology or distribution.
\subsection{Task Definition and Few-Shot Segmentation Protocol}
The 2D segmentation pipeline aims to get accurate crack segmentation masks over images from previously unseen scenarios.
Following the few-shot learning framework established by \cite{wang2019panet}, a training and testing protocol is adopted that distinguishes between seen and unseen domains.

Given a set of images for a single class \( C \), the dataset is partitioned into seen and unseen scenarios, each with associated ground truth (GT) annotations.
The training set \( D_{\text{train}} \) includes images from the seen scenarios, and the test set \( D_{\text{test}} \) includes images from the unseen scenarios. The segmentation model \( M \) is trained on \( D_{\text{train}} \) and evaluated on \( D_{\text{test}} \) ($D_{\mathrm{test}}\cap D_{\mathrm{train}}=\emptyset$).

During the test phase for \( K \)-shot supervised segmentation, \(K\) samples are randomly drawn from the unseen scenarios.  These samples include \( K \) \(\langle\text{image}, \text{mask}\rangle\) pairs representing \( K \) different variations or instances of class \( C \).
The model first extracts knowledge from these samples and then applies this knowledge to perform segmentation on the rest of the data \( D_{\text{test}} \). After training, the segmentation model \( M \) is evaluated for its \( K \)-shot segmentation performance on the test set \( D_{\text{test}} \).

In this study, the model was pre-trained using publicly available datasets. Evaluations are performed under 0-shot, 10-shot, and 110-shot configurations to assess generalization performance in our dataset.
\begin{algorithm}[tb]
\caption{Perform SAM Inference}
\label{alg:sam_inference} 
\begin{algorithmic}[1]
\Procedure{Inference}{$image$}
    \State \textbf{Input:} $image$
    \Comment{raw image}
    \State \textbf{Output:} $finalResults$ \Comment{Refined segmentation results}
    \State $mask \gets \text{DeeplabSegmentation}(image)$
    
    \State \text{Crop $image$ and corresponding $mask$ into } $PairedBatches$
    \For{$batch$ in $PairedBatches$}
        \State $mask_{crop}, image_{crop} \gets batch$
        \State $skeleton \gets \text{EuclideanDistanceTransform}(mask_{crop})$
     \State $keyPoints \gets \text{TopKPoints}(skeleton)$
    \State $sampledPoints \gets \text{DistanceBasedSampling}(keyPoints)$
        \State $promptPoints \gets \text{GeneratePrompts}(sampledPoints)$
        \State $refinedmask_{crop} \gets \text{ApplySAM}(image_{crop}, promptPoints)$
        \State \textbf{Append} $refinedmask_{crop}$ \textbf{to} $refinedResults$
    \EndFor
    
    \State $qualityScores \gets \text{AssessQuality}(refinedResults)$
    \State $finalResults \gets \text{RejectPoorResults}(refinedResults, qualityScores)$
    
    \State \Return $finalResults$
\EndProcedure
\end{algorithmic}
\end{algorithm}

\subsection{Segmentation Refinement via SAM}\label{Refine Process}
\begin{figure*}[!htbp]
    \centering
    \includegraphics[width=0.7\linewidth]{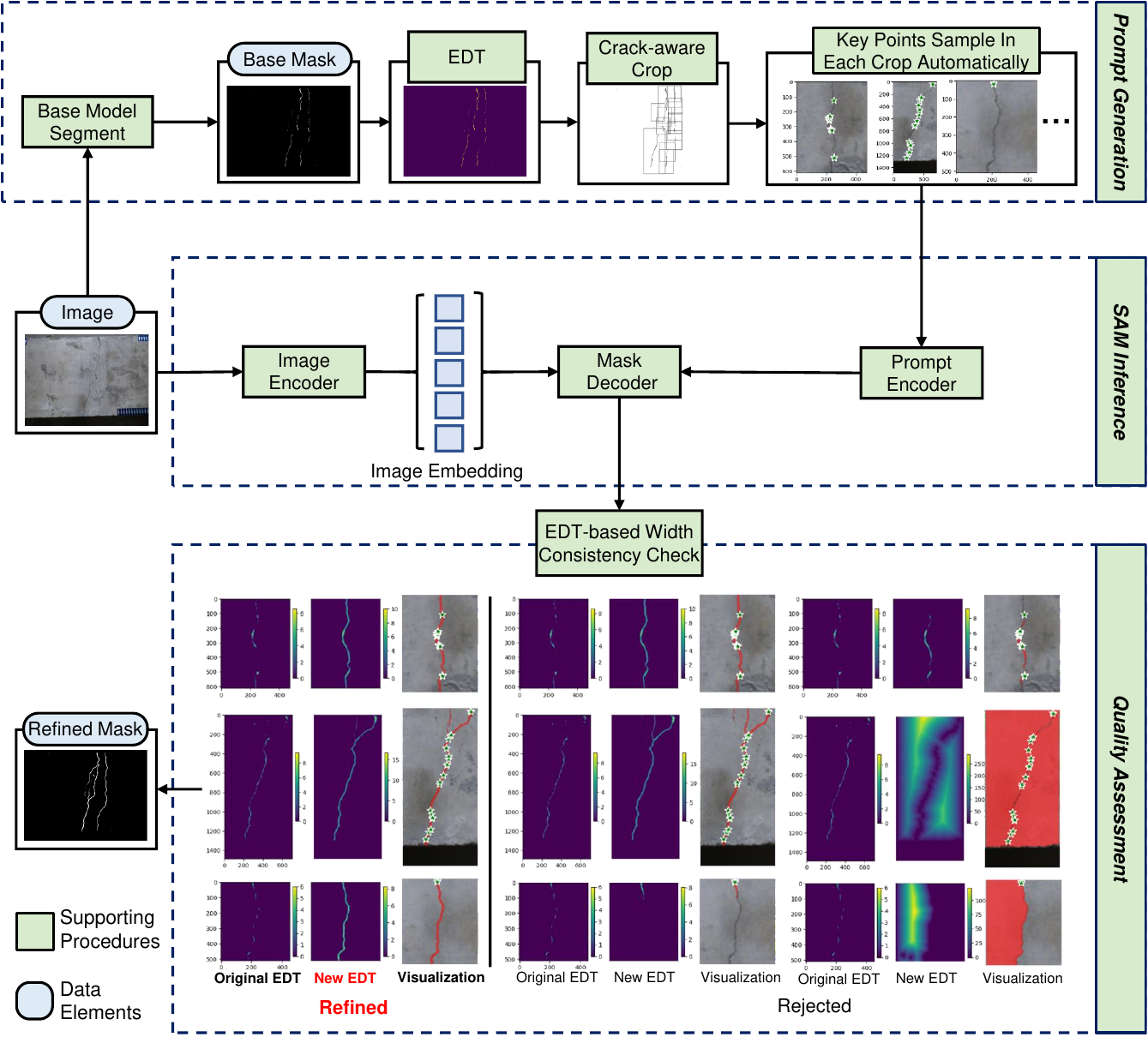}
\caption{Overview of the proposed segmentation refinement via SAM framework, composed of three stages: prompt generation, SAM inference, and quality assessment.}
    \label{fig:SAM_Framework}
\end{figure*}

Even though previous crack segmentation methods have shown effectiveness in controlled scenarios, they often struggle under unseen conditions and varying environmental factors. This challenge, known as domain shift, arises from changes in camera intrinsic parameters, hardware properties, and environmental factors such as semantic context and illumination.
Recent foundation models built on large-scale datasets like SAM \citep{kirillov2023segment} have shown zero-shot performance across different segmentation tasks. However, when tested on concrete cracks, they struggle to achieve the expected results because cracks often conflict with salient objects in the scene and typically appear at the edges of these objects.

In this framework, the base segmentation model is not expected to provide complete or final crack masks.
Its primary role is to generate coarse and possibly fragmented crack cues, which serve as spatial prompts to activate the global reasoning capability of SAM.
A 2D crack segmentation refinement pipeline-illustrated in Figure~\ref{fig:SAM_Framework} and detailed in Algorithm~\ref{alg:sam_inference}-is proposed to address the generalization issues. The pipeline refined segmentation masks by converting imperfect outputs from the base model into point prompts, which are then processed by foundation model SAM. A quality assessment module ensures the reliability of the refined masks. The process includes two core components: crack-aware prompt generation and quality assessment, as outlined in Sections \ref{prompt generation} and \ref{quality assessment}.

\subsubsection{Crack-aware Prompt Generation}\label{prompt generation}
Existing segmentation networks such as DeepLabv3+ \citep{deeplab3+} often produce masks that, while achieving high accuracy (95\%) on seen data, become incomplete and fragmented in novel scenarios.
Importantly, in real-world crack inspection, performance degradation in novel scenes is rarely caused by complete missed detection.
Instead, the dominant failure mode lies in incomplete and fragmented crack segmentation, where cracks are detected as sparse or discontinuous responses due to domain shifts in surface appearance, illumination, and background clutter.
Even under such conditions, these partial crack responses usually preserve sufficient spatial and topological cues to indicate crack existence and continuity, and can therefore be exploited as effective prompts for subsequent refinement.
Therefore, these base masks segmented by the base model can serve as high-quality prompts for the foundation model SAM.
In this work, these base masks are converted into SAM point prompts through a two-step automated process: first, its skeleton is extracted using a Euclidean Distance Transform (EDT), which generates a distance map; second, distance-based sampling on this map is performed to select the top-k points automatically.
To minimize noise and false positives, clustering algorithms are applied to group spatially close prompt points referred to as key areas. The raw image is then automatically cropped into smaller regions centered around key areas, with dilation used to ensure coverage of critical regions.
These cropped images are encoded into image embeddings via SAM's image encoder, and the above sampled top-k points are converted into prompt input using SAM's prompt encoder.
During inference, SAM's mask decoder generates refined masks from the image and prompt embeddings.
The process of generating prompt embeddings is fully automated. No manual input is required from the operator. 

\subsubsection{Quality Assessment}\label{quality assessment}

Due to the inherent randomness of prompt sampling and the semantic complexity of real-world scenes, the masks produced by SAM may occasionally exhibit unstable behaviors, most notably excessive thickening or leakage into non-crack regions.
To mitigate this issue, a lightweight quality assessment and rejection mechanism is introduced to filter unreliable SAM-refined masks using crack-specific geometric priors.
In the current implementation, the quality assessment focuses on a width consistency constraint derived from the EDT.

For each cropped region, the EDT is computed for both the baseline crack mask and the SAM-refined mask.
Let $d_{\max}^{\mathrm{SAM}}$ and $d_{\max}^{\mathrm{base}}$ denote the maximum EDT values of the SAM-refined mask and the baseline mask within the same cropped region, respectively.
Since the EDT maximum is proportional to the local crack width scale, this measure provides a robust geometric prior for detecting abnormal mask expansion.
The SAM-refined mask is rejected if
\begin{equation}
d_{\max}^{\mathrm{SAM}} > \tau_w \, d_{\max}^{\mathrm{base}}
\end{equation}
where $\tau_w = 2$ is empirically determined and fixed across all experiments.
This constraint effectively suppresses failure cases where SAM produces overly thick or blob-like regions that are inconsistent with the baseline crack geometry.

Next, a contour detection algorithm is to extract all contours \( C = \{c_1, c_2, \ldots, c_n\} \). These contours are analyzed hierarchically: a contour \( c_i \) is identified as a hole if it is nested within another contour, i.e., it has a valid parent in the contour hierarchy. Let \( P(c_i) \) denote the \emph{parent index (ID)} of contour \( c_i \), where \( P(c_i) = -1 \) indicates that \( c_i \) has no parent. The total number of holes \( H \) in the mask is then calculated as follows:
\begin{equation}
    H = \sum_{i=1}^{n} \mathbb{I}\big(P(c_i) \neq -1\big)
\end{equation}
where \( \mathbb{I}(\cdot) \) denotes the indicator function, which returns 1 if the condition is true and 0 otherwise.
These quality metrics are subsequently used to assess whether the segmentation results should be retained or discarded. This process helps ensure that only topologically and geometrically consistent masks are used to refine the final segmentation output.
If a SAM-refined mask is rejected, it is not removed from evaluation and is not replaced by an empty mask.
Instead, the system falls back to the corresponding baseline prediction for that region.
As a result, each test image always contributes exactly one final segmentation result.
This design ensures fair comparison across methods and prevents artificial performance degradation due to rejection.

\section{3D Semantic and Colorized Crack Model Reconstruction and Automatic Measurement}\label{sec6}
Current methods remain unable to simultaneously characterize crack width and spatial position due to a core challenge: the lack of high-precision fusion between crack semantic information and 3D geometric data. Recent multi-modal SLAM-based approaches attempt to address this by aligning semantic cues from Visual-Inertial Odometry (VIO) with geometric data from LiDAR-Inertial Odometry (LIO) to generate colorized point cloud~\citep{feng_crack_2023}. While this method enables high-accuracy structural reconstruction, the integration of crack semantics into the point cloud remains suboptimal. This is primarily because the fusion process heavily relies on high-frame-rate visual input, and the performance of VIO degrades significantly at lower frame rates.
In practical scenarios, capturing sub-millimeter cracks-particularly from medium or long distances-requires high-resolution cameras. However, due to bandwidth and hardware constraints, such cameras typically operate at low frame rates, making it difficult for VIO to maintain performance on par with LIO. This mismatch introduces significant temporal synchronization errors and spatial misalignment between the visual and LiDAR modules, often manifesting as localized semantic distortions in the point cloud. For sub-millimeter cracks, even minor projection or registration inaccuracies can result in severe semantic ambiguity, including blurred boundaries and distorted contours.

To address these challenges while maintaining compatibility with low-frame-rate, high-resolution cameras, a crack-oriented, multi-frame multi-modal fusion framework is proposed. This method utilizes precisely estimated camera poses from LIO and calibrated sensor extrinsic to achieve accurate alignment between image pixels and 3D point cloud, enabling robust fusion of crack semantics into the reconstructed point cloud.

This section presents a framework for high-precision 3D crack measurement, comprising four core components: multi-sensor calibration, structure reconstruction, multi-frame multi-modal fusion, and automated geometric measurement. The proposed method enables accurate fusion of multi-frame multi-modal data, aligning image-based crack semantic masks with LiDAR-generated structural geometry. Additionally, the system supports fully automated extraction of crack properties, including width and 3D position, which are essential for quantitative assessment and structural diagnostics.

\subsection{Multi-sensor Calibration}\label{sec4}
Accurate extrinsic calibration is critical for reliable data fusion among LiDAR, camera, and Inertial Measurement Unit (IMU) sensors. For both camera-IMU and camera-LiDAR setups, the goal is to estimate the rigid transformation-rotation and translation-between sensor frames.

Camera-IMU calibration is conducted using the Kalibr toolbox~\citep{6225005,6696514}, which applies full-batch optimization with spline-based motion modeling to jointly estimate temporal pose and extrinsic parameters.

For camera-LiDAR calibration, the objective is to compute the transformation \( \mathbf{T}_L^C = [\mathbf{R}_L^C \;|\; \mathbf{t}_L^C] \), where \( \mathbf{R}_L^C \in \mathrm{SO}(3) \) and \( \mathbf{t}_L^C \in \mathbb{R}^3 \). A LiDAR point \( \mathbf{P}_L \) is transformed to the camera frame as:
\begin{eqnarray}
\mathbf{P}_C = \mathbf{R}_L^C \mathbf{P}_L + \mathbf{t}_L^C
\label{eq:transform3dPts}
\end{eqnarray}
and projected onto the image plane via the intrinsic matrix \( \mathbf{K} \) as:
\begin{eqnarray}
\mathbf{P}_L^{uv} = \mathbf{K} \left[ \left( \mathbf{R}_L^C \mathbf{P}_L + \mathbf{t}_L^C \right) / Z_C \right]
\end{eqnarray}

\begin{figure}[tb]
    \centering
    \includegraphics[width=1\linewidth]{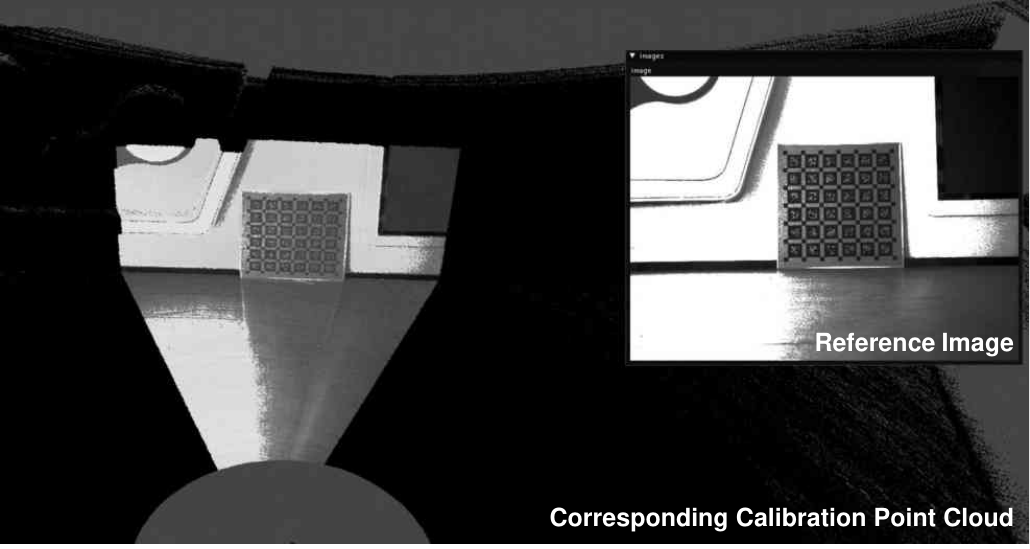}
   \caption{Calibration results show the alignment between the reference image and the corresponding LiDAR point cloud using a chessboard target.}
    \label{fig:CALIBRATION}
\end{figure}

To enhance calibration robustness, the Direct Visual-LiDAR (DVL) method~\citep{koide2023general} is used, which consists of end-to-end keypoint detection and Graph Matching Networks~\citep{li2019graphmatchingnetworkslearning}. A dense point cloud is first reconstructed to support geometric alignment between LiDAR and image data. Initial extrinsic estimates are obtained via SuperGlue~\citep{Sarlin_2020_CVPR}, which establishes keypoint correspondences between LiDAR intensity maps and images using attention-based graph matching. To refine these estimates, the Normalized Information Distance (NID)~\citep{6224750} is minimized, improving alignment by maximizing mutual information between the two sensor modalities. Final optimization is performed using the Nelder-Mead method~\citep{10.1093/comjnl/7.4.308}, yielding accurate and robust extrinsic calibration results as shown in Figure~\ref{fig:CALIBRATION}, which illustrate the spatial alignment between the camera and LiDAR sensors.

\subsection{3D Structure Reconstruction and Denoise}\label{sec6.1}
\label{LiDAR-IMU Odometry and Reconstruction}
To enhance structure reconstruction accuracy, several improvements have been implemented.

First, FastLIO2~\citep{9697912}, a state-of-the-art LiDAR-Inertial Odometry (LIO) framework that combines an iterative Kalman filter with an incremental k-dimensional tree (IKD-Tree) for real-time and high-fidelity mapping is adopted to generate dense point cloud and LiDAR poses. The first IMU frame is designated as the global frame \(G\), and the extrinsic transformation between LiDAR and IMU is defined as \(\prescript{I}{}{T}_{L} = (\prescript{I}{}{R}_{L}, \prescript{I}{}{t}_{L})\), where \(\prescript{I}{}{R}_{L} \in \mathrm{SO}(3)\) and \(\prescript{I}{}{t}_{L} \in \mathbb{R}^3\) represent rotation and translation.
The IMU motion is modeled using the following continuous-time state transition equations:
\begin{eqnarray}
\begin{aligned}
    \prescript{G}{}{\dot{R}}_{I} &= \prescript{G}{}{R}_{I} [\omega_m - b_{\omega} - n_{\omega}]_{\wedge}, \quad
    \prescript{G}{}{\dot{p}}_{I} = \prescript{G}{}{v}_{I} \\
    \prescript{G}{}{\dot{v}}_{I} &= \prescript{G}{}{R}_{I}(a_m - b_a - n_a) + \prescript{G}{}{g}
\end{aligned}
\label{eq:state-transition}
\end{eqnarray}

\noindent where $\prescript{G}{}{R}_{I} \in \mathrm{SO}(3)$ and $\prescript{G}{}{p}_{I} \in \mathbb{R}^3$ denote the IMU orientation and position in the global frame; $\prescript{G}{}{v}_{I} \in \mathbb{R}^3$ is the velocity; $a_m$, $\omega_m \in \mathbb{R}^3$ are the measured acceleration and angular velocity; $b_a$, $b_{\omega} \in \mathbb{R}^3$ are the corresponding IMU biases, and $n_a$, $n_{\omega} \in \mathbb{R}^3$ represent zero-mean Gaussian noise. $\prescript{G}{}{g} \in \mathbb{R}^3$ is the gravity vector, and $[\cdot]_{\wedge}$ denotes the skew-symmetric operator mapping $\mathbb{R}^3$ to $\mathfrak{so}(3)$.

FastLIO2 uses the IKD-Tree~\citep{9697912} to incrementally manage large-scale LiDAR data, enabling efficient map construction without sacrificing accuracy. Its integration with Kalman-based state estimation ensures robust and consistent LiDAR pose estimation essential for dense reconstruction.

Furthermore, to recover camera poses aligned with LiDAR frames, timestamp synchronization is first performed. The camera's pose in \(\mathrm{SE}(3)\) is then interpolated from adjacent LiDAR poses using spherical linear interpolation (Slerp). Given two poses \(p_0, p_1 \in \mathrm{SE}(3)\) and a normalized time parameter \(t \in [0, 1]\), the interpolated pose is given by:
\begin{eqnarray}
\text{slerp}(p_0, p_1; t) = \frac{\sin[(1 - t)\Omega]}{\sin \Omega} p_0 + \frac{\sin[t\Omega]}{\sin \Omega} p_1
\end{eqnarray}
where \(\Omega = \cos^{-1}(p_0 \cdot p_1)\) is the angle between the unit quaternions representing the poses. This method provides smooth, continuous camera trajectories aligned with the LiDAR-based reconstruction, supporting precise texture projection and crack localization.

Then, to mitigate the noise in the point cloud, a two-step approach is proposed, consisting of denoising and smoothing. First, a Statistical Outlier Removal (SOR) filter from the Point Cloud Library (PCL) is used to remove outliers based on the distance distribution to $K$-nearest neighbors, using a sigma-based threshold~\citep{balta_fast_2018}:
\begin{eqnarray}
r_i(K) > \mu_r + N \cdot \sigma_r
\end{eqnarray}
where $r_i(K)$ is the mean distance to neighbors, and $\mu_r$, $\sigma_r$ are the global mean and standard deviation. Typical settings ($K=60$, $N=1$) ensure 68\% inlier retention under normality assumptions.

Next, the Moving Least Squares (MLS) method~\citep{10.1007/11553595_88} is employed to reconstruct smooth surfaces from unstructured point samples. Given a set of points $S = \{(x_i, f_i)\}$, MLS fits a polynomial $\tilde{p}(x)$ of degree $m$ by minimizing:
\begin{eqnarray}
\tilde{p}(x) = \arg \min_p \sum_i [p(x_i) - f_i]^2 \theta(\|x - x_i\|)
\end{eqnarray}
where $\theta(s) = e^{-s^2}$ gives higher weights to nearby samples. The smoothed surface is used to project original points, producing a refined and coherent point cloud.

This pipeline enables extraction of clean structural regions and supports downstream tasks such as coloring, crack segmentation, and 3D geometry computation.
\begin{figure*}[htbp]
    \centering
    \includegraphics[width=0.95\linewidth]{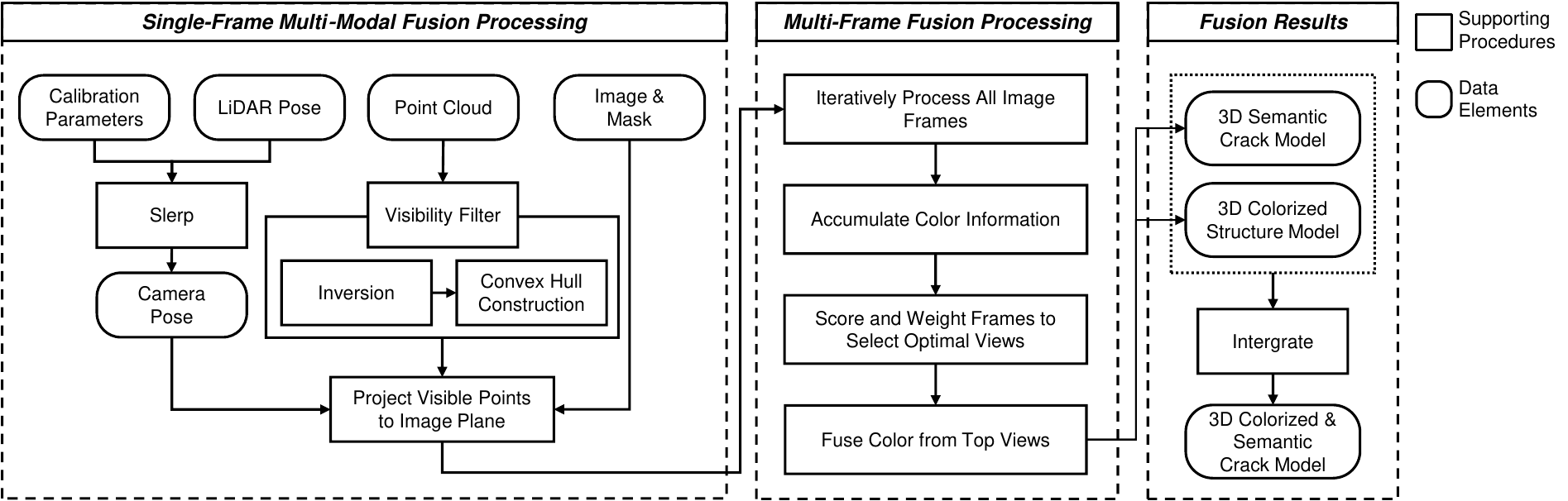}
\caption{Workflow of the proposed crack-oriented multi-frame and multi-modal fusion, integrating image and LiDAR data to generate a 3D colorized \& semantic crack model.}
    \label{fig:fusion}
\end{figure*}

\begin{figure*}[htbp]
    \centering
    \includegraphics[width=0.85\linewidth]{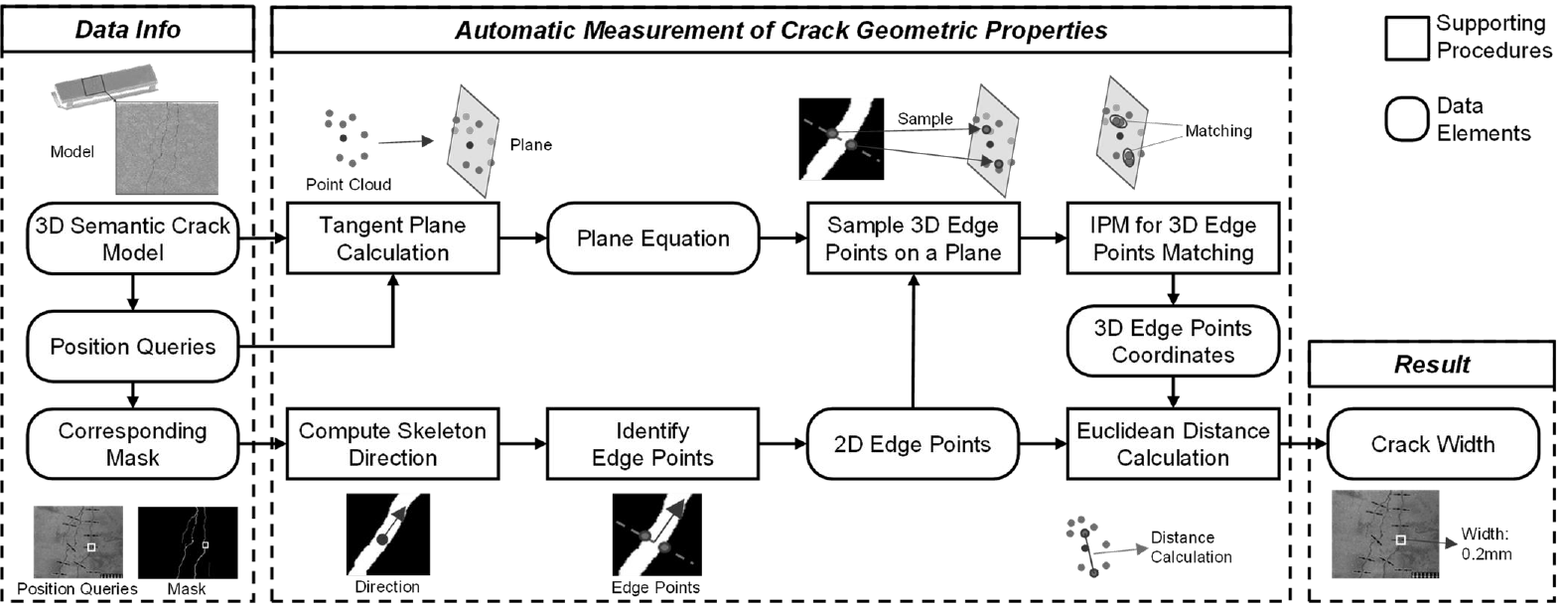}
\caption{Workflow for automatic measurement of crack geometric properties from the 3D semantic crack model, including spatial localization and width estimation.}
    \label{fig:width}
\end{figure*}

\subsection{Crack-Oriented Multi-Frame Multi-Modal Fusion}\label{sec7.1}
To achieve accurate fusion of crack semantic information from multiple image frames into a coherent 3D structural point cloud, the core challenge in this process is to precisely align between 2D image pixels and 3D point cloud. To overcome this, a crack-oriented, multi-frame multi-modal fusion framework is proposed, as illustrated in Figure~\ref{fig:fusion}.
The overall pipeline in Figure~\ref{fig:fusion} consists of (i) standard geometric processing blocks commonly used in multi-view point-cloud projection (pose interpolation, visibility filtering, and perspective projection), and (ii) a crack-oriented view scoring and weighting strategy proposed in this study to select the most informative frames for crack-aware color and semantic fusion. The latter is specifically designed to favor observations that best preserve fine crack details under practical inspection conditions.

The fusion process begins by converting the 3D LiDAR point cloud into the camera coordinate system for each image frame. This transformation leverages the camera pose estimated through Slerp in the \textit{SE}(3) Lie group \citep{pennec1998computing}, along with the extrinsic calibration parameters of camera-LiDAR.
This ensures spatial alignment of the LiDAR points with the corresponding camera perspective.
To identify which points are visible from the current camera pose, the Hidden Point Removal (HPR) operator \citep{katz2007direct} is utilized. This visibility determination involves two key steps: a spherical inversion followed by convex hull construction. For each point \(\mathbf{P}_i \in P\), an inversion is applied relative to a sphere of radius \(R\) centered at the camera position \(\mathbf{C}\), expressed as:
\begin{equation}
    \hat{\mathbf{P}}_i = \mathbf{P}_i + 2\bigl(R - \|\mathbf{P}_i-\mathbf{C}\|\bigr)\frac{\mathbf{P}_i-\mathbf{C}}{\|\mathbf{P}_i-\mathbf{C}\|}
    \label{eq:HprOperator}
\end{equation}
where \(\|\mathbf{P}_i-\mathbf{C}\|\) denotes the Euclidean norm of the vector \(\mathbf{P}_i-\mathbf{C}\), i.e., the Euclidean distance from \(\mathbf{P}_i\) to the camera center \(\mathbf{C}\).

The set of inverted points \(\hat{P} = \{\hat{\mathbf{P}}_i\} \cup \{C\}\) is then used to compute the convex hull. A point is marked as visible if its inverted counterpart lies on this convex hull, effectively filtering out occluded points from the camera’s view.
The resulting visible points, denoted in the camera coordinate frame as \(\mathbf{P}_C = [X_C, Y_C, Z_C]^T\), are projected onto the 2D image plane using the camera intrinsic matrix \(\mathbf{K}\):
\begin{eqnarray}
     \mathbf{p}_{uv} = \mathbf{K} \begin{bmatrix} {X_C}/{Z_C} & {Y_C}/{Z_C} & 1 \end{bmatrix}^T
\end{eqnarray}
where \(\mathbf{p}_{uv} = [u, v]^T\) are the pixel coordinates corresponding to the projected 3D point.
Color information from each image frame is then assigned to the projected 3D points.

As the fusion proceeds across all frames, each point aggregates multiple color observations. 
To determine the most suitable image sources for crack-aware fusion, a crack-oriented frame weighting scheme is designed. The approach prioritizes views with (i) favorable viewing direction and (ii) an appropriate observation range, which empirically helps preserve thin crack textures and reduces background contamination in the fused crack semantics.
The first score assesses the angular alignment between the point-to-camera vector
\(\mathbf{D}_{\text{ptc},i}\) and the camera optical axis \(\mathbf{N}_{\text{cpn},i}\)
in the \(i\)-th frame, reflecting view orientation:
\begin{equation}
\text{Score}_{1,i} = \cos(\theta_i) =
\frac{\mathbf{D}_{\text{ptc},i} \cdot \mathbf{N}_{\text{cpn},i}}
{\|\mathbf{D}_{\text{ptc},i}\| \|\mathbf{N}_{\text{cpn},i}\|}
\end{equation}
The second score evaluates the spatial proximity of the point to the camera in the
\(i\)-th frame, favoring observations at an ideal range (e.g., 2 meters):
\begin{equation}
\text{Score}_{2,i} = \exp\left(-\frac{(\|\mathbf{D}_{\text{ptc},i}\| - 2)^2}{2\sigma^2}\right)
\end{equation}
where \(\sigma\) is a user-defined parameter controlling the decay rate with respect to distance deviation.
The final weight assigned to the \(i\)-th frame is calculated as a linear combination of the two scores:
\begin{equation}
w_i = \lambda_1 \cdot \text{Score}_{1,i} + \lambda_2 \cdot \text{Score}_{2,i}
\end{equation}
with \(\lambda_1\) and \(\lambda_2\) controlling the balance between angular and distance preferences.
For each 3D point, the top \(N\) frames with the highest weights are selected. 
In this study, the weighting parameters are set empirically.
Specifically, equal weights (\(\lambda_1=\lambda_2\)) are assigned based on preliminary experiments across different inspection scenes, which were found to provide stable and visually consistent fusion results.
This choice reflects that view orientation and observation distance play similarly important roles in ensuring reliable color assignment and semantic projection for crack-related points.
It is also observed that the fusion results are not highly sensitive to moderate variations of \(\lambda_1\) and \(\lambda_2\).
This robustness is mainly because the weighting strategy only affects the relative ranking of candidate views, and small changes in the weights rarely alter the selected top \(N\) frames.

The final color is then computed as a weighted average of RGB values from these selected frames:
\begin{eqnarray}
    \text{Color}_{\text{final}} = \frac{\sum_{i=1}^{N} w_i \cdot \text{Color}_i}{\sum_{i=1}^{N} w_i}
\end{eqnarray}
where \(\text{Color}_i\) denotes the RGB value retrieved from the \(i\)-th image.
This fusion mechanism ensures that color assignment originates from geometrically optimal views, thereby enhancing both the visual fidelity and semantic clarity of the reconstructed point cloud.
By aligning multi-frame image data with the reconstructed geometry, a 3D colorized structural model is generated. Simultaneously, incorporating segmentation masks enables the construction of a high-precision 3D semantic crack model.
The integration of these two components results in a unified 3D colorized and semantic crack model, enabling accurate restoration of surface textures and detailed crack characterization, and providing a robust foundation for automated geometric measurement.

\subsection{Automatic Measurement of Crack Geometric Properties}\label{sec7.2}
After obtaining a 3D point cloud enriched with semantic crack labels and real-world scale, the next step is the automatic measurement of crack width and spatial location. While direct manual measurement by selecting crack edge points is possible, this approach is subjective and time-consuming, especially when evaluating numerous cracks.
More importantly, the reconstructed point cloud has a precision at the millimeter level, and the uncertainty caused by its discretization introduces measurement errors, which cannot guarantee that crack measurements meet millimeter-level precision.
To overcome these limitations, as illustrated in Figure~\ref{fig:width}, a novel fully automated measurement pipeline is developed. The proposed framework extracts crack skeletons and edge points from 2D segmentation masks and maps them back to 3D space using calibrated camera parameters. A local crack surface is estimated through surface fitting, and the projected edge information is used to determine accurate 3D coordinates for width measurement. The final crack width is defined as the Euclidean distance between the corresponding 3D edge points.

The process starts with estimating the direction of the crack skeleton. A local neighborhood around each skeleton point is selected and smoothed using a Gaussian filter. The Sobel operator is then applied to compute local image gradients, and the averaged gradient yields the crack direction vector, which is normalized and used for subsequent edge tracing.

Next, the direction vector and its perpendicular direction are employed to iteratively trace and identify the left and right crack edge points in the 2D segmentation mask. To improve the accuracy of 3D edge localization, a local surface plane is estimated by least-squares fitting to a small neighborhood of the reconstructed 3D semantic point cloud around the crack measurement location.
The fitted plane is represented as
\begin{equation}
a x + b y + c z + d = 0
\end{equation}
where $\mathbf{n}=[a,b,c]^{\top}$ denotes the plane normal vector. Two orthonormal basis vectors $\mathbf{u}$ and $\mathbf{v}$ are then constructed such that $\mathbf{u}\perp\mathbf{n}$, $\mathbf{v}\perp\mathbf{n}$, and $\mathbf{u}\perp\mathbf{v}$. Using the neighborhood center $\mathbf{p}_0$, which corresponds to the crack measurement location, candidate points on the fitted plane are generated through a local 2D parameterization:
\begin{equation}
\mathbf{p}(s,t)=\mathbf{p}_0+s\,\mathbf{u}+t\,\mathbf{v}
\end{equation}
where $\mathbf{u}$ and $\mathbf{v}$ span the local plane.

These sampled points form a constrained candidate set for crack edge localization. Using the known camera poses and intrinsic parameters, the 2D crack boundary points are mapped into 3D space via inverse perspective mapping (IPM) \citep{BERTOZZ1998585}, yielding 3D coordinates constrained to the fitted plane. 
The sampled candidate points are also projected onto the fitted plane, and the optimal matching point is identified based on a minimum-error criterion between the projected crack boundary points and the candidate set. 
The 3D coordinates of the crack edge points are thus determined from the matched candidate points, and the crack width is computed as the Euclidean distance between the left and right edge points.

The proposed automated framework provides a reliable and repeatable solution for simultaneously extracting crack width and 3D positional information from semantically segmented point clouds. This significantly improves processing efficiency and enables comprehensive crack analysis and quantitative evaluation.

\section{Experiment Setup}\label{sec8}
\subsection{Platforms}\label{sec8.1}
\begin{figure}[bt]
    \centering
    \includegraphics[width=0.85\linewidth]{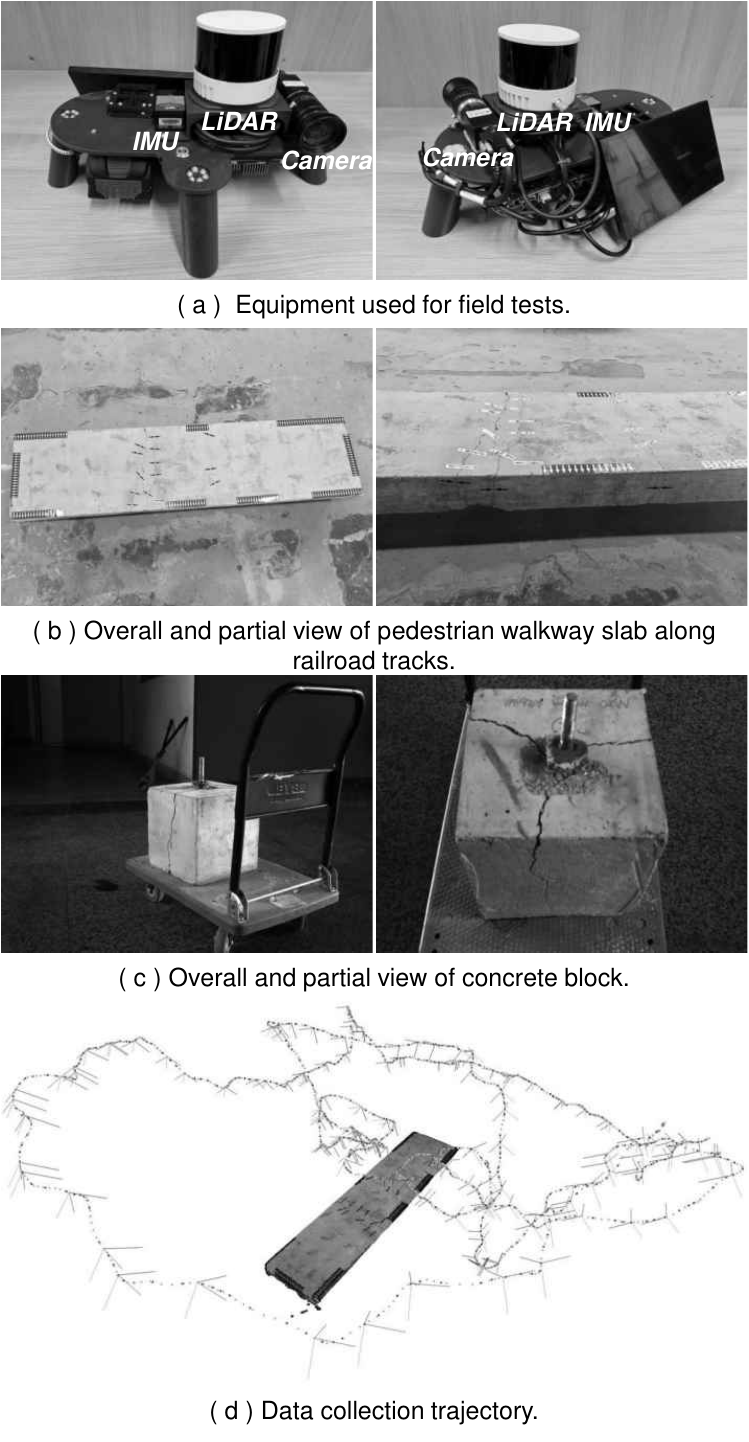}
    \caption{Overview of experimental setup and test specimens used in the study.}
    \label{fig:hardware}
\end{figure}

This study presents a state-of-the-art 3D reconstruction system based on SLAM technology, specifically designed for crack detection and measurement in concrete structures.
By seamlessly integrating LiDAR point cloud and image data, the system significantly enhances data accuracy, enabling precise crack measurements and efficient real-time 3D reconstruction.
To facilitate real-world applications, a handheld device (Figure~\ref{fig:hardware}a) equipped with a global shutter camera, lens, LiDAR, and IMU was developed, as detailed in Table~\ref{tab:equipment_parameters}. 

The LiDAR captures precise point-cloud data while the camera is utilized for high-resolution image acquisition. 
The IMU records the high-frequency motion changes of the device, and the industrial computer coordinates system operations and stores the collected data. 
The device is powered by an Intel i7-1265UE CPU with 32 GB of RAM and an integrated power supply.
This integration of components ensures both precision in crack quantification and the robustness of the 3D reconstruction process.

\begin{table}[tb]
\centering
\caption{Specifications of the primary sensors used in the experiments.}
\label{tab:equipment_parameters}
\begin{tabular*}{240pt}{@{\extracolsep\fill}lll@{\extracolsep\fill}}%
\toprule
\textbf{Sensor} & \textbf{Type} & \textbf{Specification} \\
\midrule
LiDAR & 
\begin{tabular}[t]{@{}l@{}}
Hesai \\
XT32-M2X
\end{tabular} & 
\begin{tabular}[t]{@{}l@{}}
FOV: 360\textdegree{} $\times$ 40.3\textdegree{} \\
Angular Res.: 0.18\textdegree{} $\times$ 1.3\textdegree{} \\
Ranging Accuracy: 0.5 cm (1$\sigma$) \\
Range: 0.5–300 m
\end{tabular} \\
Camera & 
\begin{tabular}[t]{@{}l@{}}
HikRobot \\
MV-CH120-10GC
\end{tabular} & 
Resolution: 4096 $\times$ 3000 \\
Lens & 
\begin{tabular}[t]{@{}l@{}}
HikRobot \\
MVL-KF1628M-12MPE
\end{tabular} & 
Focal Length: 16 mm \\
IMU & 
\begin{tabular}[t]{@{}l@{}}
WitMotion \\
HWT905
\end{tabular} & 
Accuracy: 0.05\textdegree{} \\
\bottomrule
\end{tabular*}

\vspace{0.3em}
\begin{minipage}{240pt}
\footnotesize
\textbf{Notes:} “FOV" means Field of View; “Res." means Resolution.
\end{minipage}
\end{table}

\subsection{Model Implementation}
All segmentation and reconstruction tasks were executed on a workstation running Ubuntu 22.04 LTS, configured with an Intel i9-13900K CPU, 128~GB of RAM, and an NVIDIA RTX 4090 GPU with 24~GB VRAM. All deep learning models were implemented using PyTorch 2.0.0 and CUDA 11.5.119.
The segmentation pipeline followed a two-stage training strategy: pre-training followed by fine-tuning. In the pre-training phase, ten open-source datasets (listed in Table~\ref{Dataset}) were aggregated and partitioned into training and validation sets using an 8:2 split. A batch size of 32 was used, with an initial learning rate of 0.01, for 50 training iterations. The model checkpoint with the best validation performance was retained for subsequent tuning. 
For the few-shot evaluation in the unseen scenario, the manually labeled subset was split into a training pool and a fixed test set with a 7:3 ratio. The 10-shot setting randomly samples 10 images from the 110-image training pool, the 110-shot setting uses all 110 training-pool images, and the 0-shot setting uses no labeled images from the unseen scenario.

Training was conducted with a batch size of 4 and an initial learning rate of 0.001, for a total of 200 iterations. After both stages, segmentation results were further refined by SAM to improve spatial consistency and segmentation accuracy.
To enhance generalization and reduce overfitting, standard data augmentation techniques such as random horizontal and vertical flips, random rotations, and Gaussian blurring were applied during training.

The segmentation accuracy was evaluated using the Intersection over Union (IoU) of the crack, which measures the pixel-level overlap between the predicted crack mask and the ground-truth crack mask. It is defined as:
\begin{equation}
\mathrm{IoU} = \frac{|\mathcal{M}_{\text{pred}} \cap \mathcal{M}_{\text{gt}}|}{|\mathcal{M}_{\text{pred}} \cup \mathcal{M}_{\text{gt}}|}
\end{equation}
where \(\mathcal{M}_{\text{pred}}\) and \(\mathcal{M}_{\text{gt}}\) denote the sets of pixels labeled as crack in the predicted and ground-truth masks, respectively.

\subsection{Crack Segmentation Datasets}
\begin{table}[tb]
\centering
\caption{Dataset composition.}
\begin{tabular*}{240pt}{@{\extracolsep\fill}lll@{\extracolsep\fill}}%
\toprule
\textbf{Crack Dataset} & \textbf{Size} & \textbf{Resolution} \\
\midrule
Ceramic-Cracks & 100 & 256$\times$256 \\
CFD & 118 & 480$\times$320 \\
Crack500 & 50 & 400$\times$400 \\
CrackTree200 & 206 & 800$\times$600 \\
DeepCrack & 527 & 544$\times$384 \\
GAPS & 544 & 1920$\times$1080 \\
Masonry & 240 & 224$\times$224 \\
Rissbilder & 5591 & 448$\times$448 \\
Volker & 990 & 448$\times$448 \\
CCSS-DATA & 670 & 544$\times$384 \\
Our dataset & 150 & 4096$\times$3000 \\
\bottomrule
\end{tabular*}
\label{Dataset}
\end{table}

The experimental dataset used in this work comprises 10 open-source datasets, as detailed in Table~\ref{Dataset}. 
This dataset contains 9,260 RGB images of varying sizes captured in diverse environments, all annotated at the pixel level for cracks. 
The training and testing datasets for previously unseen scenarios consist of images captured using a handheld device in our experimental setup. Each image was manually annotated using the PixelAnnotationTool \citep{Breheret:2017}. They have a resolution of 4096×3000 pixels and vary in shooting angles, distances, and levels of blurriness. The dataset also includes artifacts such as reflective tags, shadows, stains, and miscellaneous background objects.

\subsection{Data Acquisition from Field Experiments}
The effectiveness of the proposed system for crack detection, localization, and width measurement was validated through experiments conducted on a concrete block and a pedestrian walkway slab along railroad tracks (reinforced concrete slab). The experimental subjects are shown in Figure~\ref{fig:hardware}b. The cracks that developed are mainly located at the center of the slab, with dimensions of 1580mm $\times$ 300mm $\times$ 80mm.
Figure~\ref{fig:hardware}c illustrates the concrete block, which is a standard cube with dimensions of 300mm $\times$ 300mm $\times$ 300mm, placed on a trolley. Cracking occurred due to bolt pull-out. 

During data collection from the pedestrian walkway slab, a self-developed handheld device, as described in Section \ref{sec8.1}, was directed at the pedestrian walkway slab. The data were recorded by rotating around the slab in a counterclockwise direction, with the image capture path depicted in Figure~\ref{fig:hardware}d.
During data recording, the LiDAR point cloud was collected at 10 Hz, IMU data at 200 Hz, and images at approximately 5 Hz. The total recording time was about 2 minutes. 
The same procedure was used to collect data from the concrete block.
Additionally, to verify the feasibility of the crack width measurement method, four major cracks were selected on the pedestrian walkway slab. These cracks traversed all four faces of the slab. 

At intervals of approximately 10 mm along the crack direction, the actual crack width was measured using a Dino-Lite AF4915 microscope. The measured values were then compared with the 3D crack width information obtained by the proposed method. 
To facilitate the matching of the actual crack measurements with the corresponding locations in the 3D reconstruction results, yellow reflective markers were placed at the measurement points, as shown in Figure~\ref{fig:hardware}b.
These markers were used only to relocate the microscope measurement points for evaluation and were not used by any step of the proposed pipeline.
They are not necessary to achieve the reported accuracy and may introduce visual artifacts that can even degrade the segmentation of baseline models.

\begin{figure*}[tb]
    \centering  
    \includegraphics[width=0.95\linewidth]{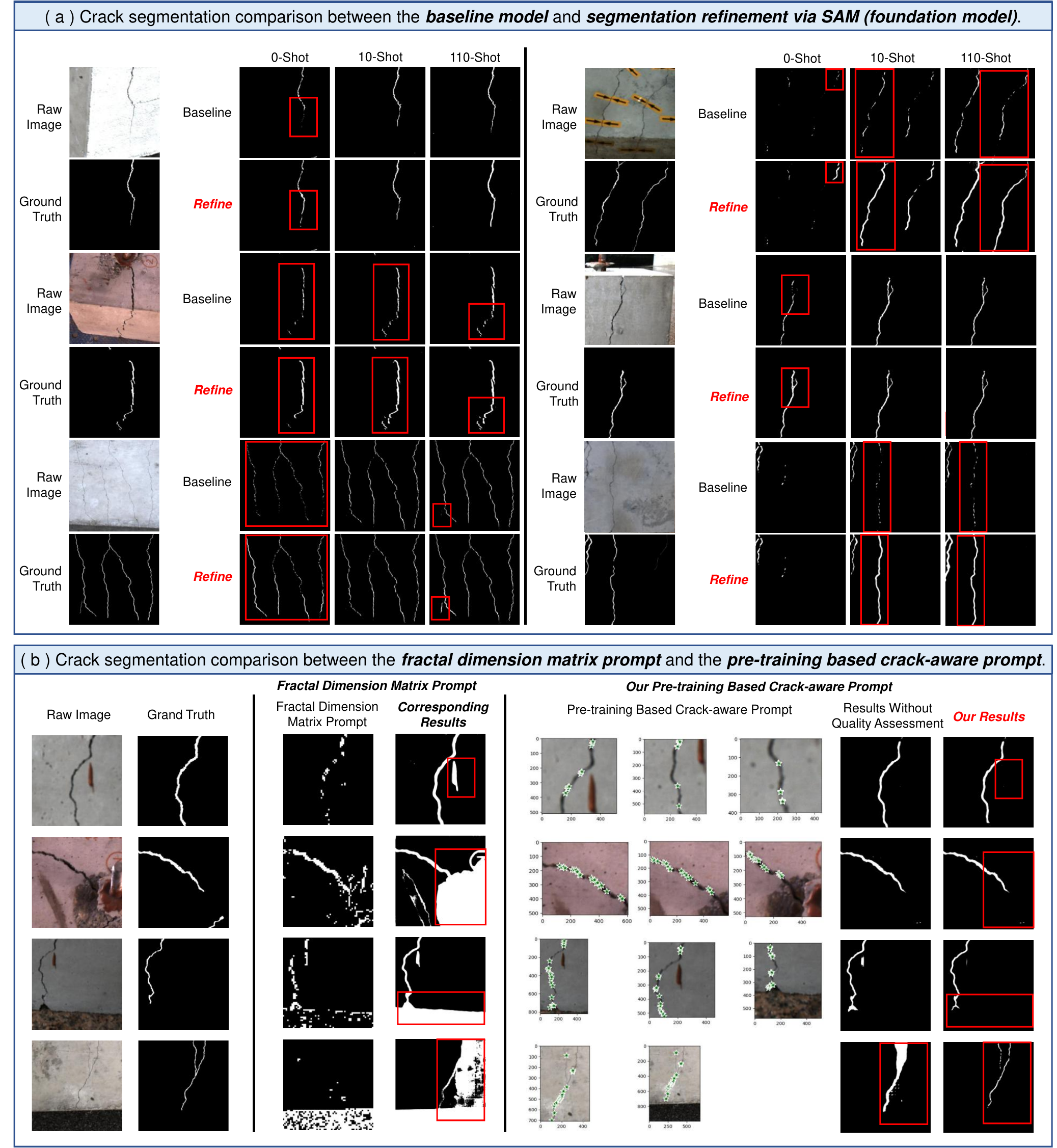}
    \caption{Examples of result comparisons for crack segmentation tasks. Red boxes indicate where our method significantly outperforms compared methods.}
    \label{fig:sam picture}
\end{figure*}

In subsequent experiments, as shown in Figure~\ref{fig:sam picture}a, the markers in the scene were kept to reflect realistic surface interference. Although they significantly affected the segmentation results of the baseline model, the SAM-based refinement maintained stable performance.
This enabled the automated crack width measurement results to retain submillimeter-level accuracy, demonstrating the robustness of the proposed method. This will be fully demonstrated in Sections \ref{sec5.1} and \ref{sec5.2}.

\begin{table}[t]
\centering
\caption{Base model segmentation and segmentation refinement via SAM under different few-shot settings.}
\label{tab:performance}
\small
\setlength{\tabcolsep}{4pt}
\renewcommand{\arraystretch}{1.08}

\begin{tabular*}{\columnwidth}{@{\extracolsep{\fill}} l l c c c @{}}
\toprule
\multirow{2}{*}{\textbf{\# Shots}} & \multirow{2}{*}{\textbf{Method}} &
\multicolumn{2}{c}{\textbf{IoU (\%)}} & \textbf{Our refined} \\
\cmidrule(lr){3-4}
& & \textbf{Baseline} & \textbf{SAM-refined} & \textbf{IoU gain (\%)} \\
\midrule
\multirow{4}{*}{0}
& Swin-Transformer & 36.77 & 44.15 & \textbf{\textit{+7.38}} \\
& TernausNet       & 42.13 & 49.92 & \textbf{\textit{+7.79}} \\
& TransUnet        & 45.28 & 52.04 & \textbf{\textit{+6.76}} \\
& DeepLabv3+       & 46.61 & 53.21 & \textbf{\textit{+6.60}} \\
\midrule
\multirow{4}{*}{10}
& Swin-Transformer & 41.29 & 45.59 & \textbf{\textit{+4.30}} \\
& TernausNet       & 45.51 & 50.83 & \textbf{\textit{+5.32}} \\
& TransUnet        & 49.77 & 53.55 & \textbf{\textit{+3.78}} \\
& DeepLabv3+       & 50.13 & 55.27 & \textbf{\textit{+5.14}} \\
\midrule
\multirow{4}{*}{110}
& Swin-Transformer & 50.01 & 54.73 & \textbf{\textit{+4.72}} \\
& TernausNet       & 57.24 & 59.18 & \textbf{\textit{+1.94}} \\
& TransUnet        & 60.38 & 61.30 & \textbf{\textit{+0.92}} \\
& DeepLabv3+       & 61.41 & 62.25 & \textbf{\textit{+0.84}} \\
\bottomrule
\end{tabular*}
\end{table}

\begin{table}[tb]
\centering
\fontsize{8pt}{12pt}\selectfont
\caption{Ablation study of proposed modules.}
\begin{tabular*}{240pt}{@{\extracolsep\fill}llll@{\extracolsep\fill}}
\toprule
\textbf{Setting}  & \textbf{Prompt} & \textbf{QA \tnote{1}} & \textbf{IoU (\%)} \\
\midrule
a & None & $\times$ & 2.0 \\
b & Fractal Dimension Matrix & $\times$ & <5.0 \\
c & Our prompt & $\times$ & 22.9 \\
d & \textbf{Our prompt} & $\checkmark$ & \textbf{53.2} \\
\bottomrule
\end{tabular*}

\vspace{0.2em}
\begin{minipage}{\textwidth}
\footnotesize
\textbf{Notes:} “QA" means quality assessment.
\end{minipage}
\footnotesize          
\label{tab:prompt_results}
\end{table}

\section{Experiment Result}
\subsection {Results of Image Crack Segmentation}\label{sec5.1}
This section first evaluates the effectiveness of the proposed segmentation refinement via SAM approach across various baseline segmentation models. Additionally, an ablation study is presented to analyze the contributions of the proposed modules, including prompt generation strategies and the segmentation quality assessment mechanism.

First, DeepLabv3+ \citep{deeplab3+}, TransUnet \citep{TernausNet}, Swin-Transformer \citep{liu2021Swin}, and TernausNet \citep{TernausNet,pantoja-rosero_topo-loss_2022} were selected as baseline models to assess the improvement brought by segmentation refinement via SAM. In practical engineering structure damage detection, crack image segmentation tasks often require the segmentation model to perform pixel-level segmentation of concrete cracks in diverse environments.
To meet these demands, the performance of the networks was evaluated and compared under zero-shot, 10-shot, and 110-shot conditions.
Table~\ref{tab:performance} summarizes the IoU scores before and after applying segmentation refinement via SAM.
The results show that segmentation refinement via SAM consistently improves performance across all baseline models. Under zero-shot conditions, an average improvement of approximately 7\% in IoU was observed, while under 10-shot conditions, an average improvement of about 5\% was achieved.
Even under 110-shot settings, refinement maintained performance gains. 

Among all models, DeepLabv3+ combined with SAM refinement achieved the highest IoU score of 62.25\%.
Figure~\ref{fig:sam picture}a illustrates representative segmentation outputs from this configuration. To improve the clarity of visual presentation, all examples shown were cropped local regions from the original full-resolution images captured by the acquisition device.
Despite the presence of challenging conditions such as yellow crack labels that negatively impact baseline performance, the refined model maintained stable segmentation. Meanwhile, SAM effectively enhances segmentation across a range of complex scenarios, including concrete surfaces with crack-like noise, overexposure, branching or inclined cracks, sub-pixel-width cracks, occlusions, and stain interference. The improvements are particularly pronounced in zero-shot and 10-shot settings.

Overall, the integration of foundation models with task-specific models results in enhanced segmentation quality, reduced reliance on labeled data, and increased generalizability to unseen environments (up to 7\% in IoU). From a practical standpoint, this reduces deployment costs and enhances the method's applicability in engineering contexts. 
Based on these findings, DeepLabv3+ was selected as the base model for subsequent field experiments.

In addition, to further evaluate the proposed modules, a zero-shot ablation study was conducted using four SAM configurations: (a) no prompt, (b) fractal dimension matrix-based prompt, (c) pre-training based crack-aware prompts without the segmentation quality assessment module, and (d) pre-training based crack-aware prompts with the proposed quality assessment module. The results are shown in Table~\ref{tab:prompt_results}.

\begin{figure*}[!tbp]
    \centering
    \includegraphics[width=0.87\linewidth]{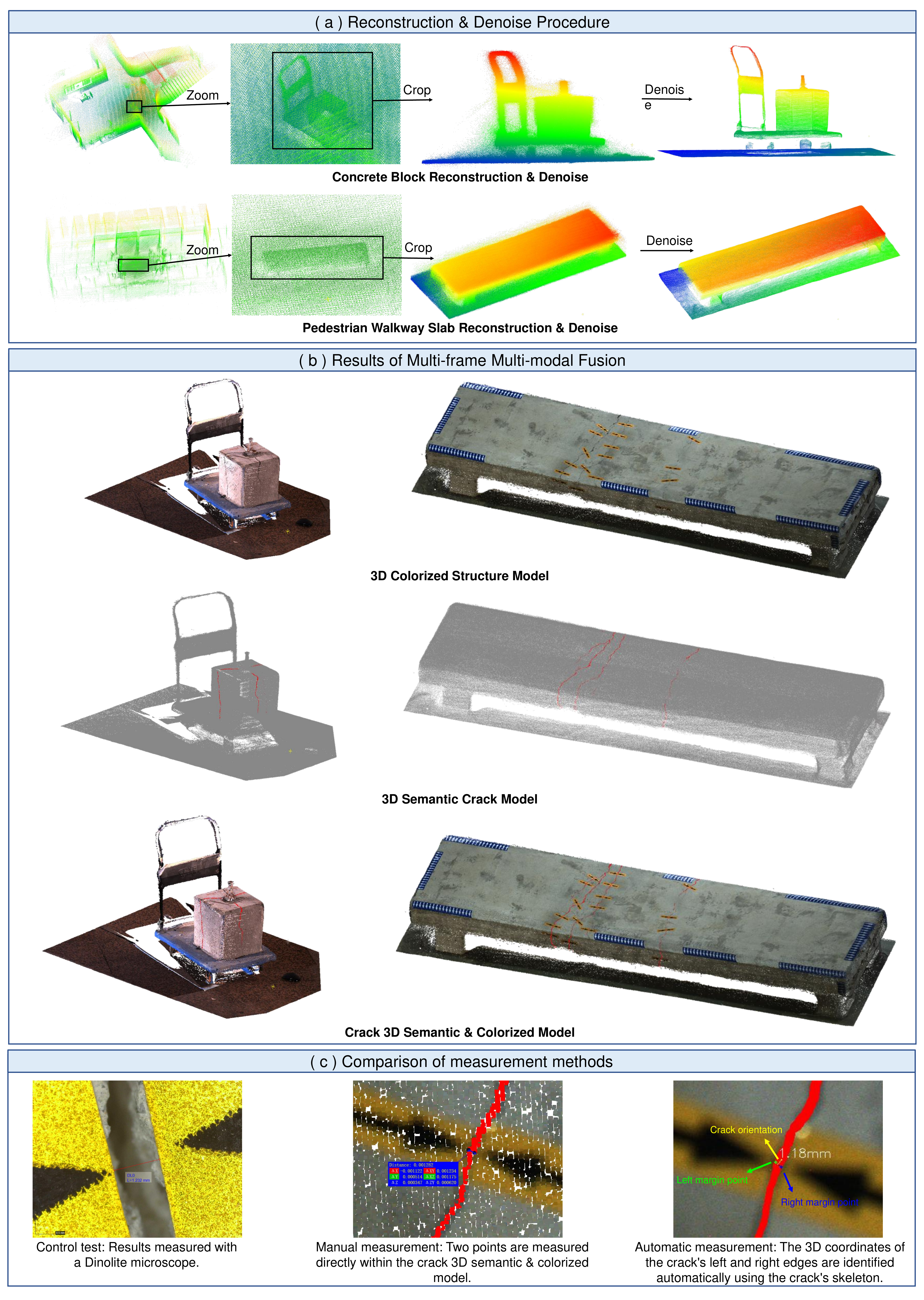}
    \caption{Schematic of structural reconstruction, crack semantic fusion and crack measurement.}
    \label{fig:reconstruction}
\end{figure*}

\begin{table}[t]
\centering
\caption{Automatic crack width measurement results.}
\label{tab:width}

\footnotesize
\sisetup{detect-all, mode=text}

\setlength{\tabcolsep}{7pt}
\renewcommand{\arraystretch}{1.10}

\begin{tabular}{@{\extracolsep{\fill}} l
S[table-format=2.2]
S[table-format=2.2]
S[table-format=2.2]
S[table-format=2.2] @{}}
\toprule
\textbf{Crack ID} &
\textbf{Calculated value} &
\textbf{Reference value} &
\textbf{AE} &
\textbf{RE} \\
(\#) & {(mm)} & {(mm)} & {(mm)} & {(\%)} \\
\midrule
\#1  & 0.59 & 0.65 & 0.06 &  9.09 \\
\#2  & 0.86 & 0.91 & 0.05 &  5.49 \\
\#3  & 0.38 & 0.56 & 0.18 & 32.26 \\
\#4  & 1.18 & 1.23 & 0.05 &  4.22 \\
\#5  & 0.96 & 0.92 & 0.04 &  4.80 \\
\#6  & 1.72 & 1.43 & 0.29 & 20.11 \\
\#7  & 0.66 & 0.67 & 0.01 &  1.49 \\
\#8  & 0.40 & 0.58 & 0.18 & 30.80 \\
\#9  & 0.30 & 0.45 & 0.15 & 33.92 \\
\#10 & 0.20 & 0.29 & 0.09 & 30.56 \\
\#11 & 0.75 & 0.86 & 0.11 & 12.28 \\
\#12 & 1.41 & 1.49 & 0.08 &  5.24 \\
\#13 & 0.32 & 0.28 & 0.04 & 15.94 \\
\#14 & 0.85 & 0.91 & 0.06 &  6.80 \\
\#15 & 0.51 & 0.52 & 0.01 &  2.49 \\
\midrule
\textbf{ME} & \multicolumn{1}{c}{--} & \multicolumn{1}{c}{--} & \textbf{0.09} & \textbf{14.37} \\
\bottomrule
\end{tabular}

\vspace{0.3em}
\begin{minipage}{\columnwidth}
\footnotesize
\textbf{Notes:} “AE” means absolute error; “RE” means relative error; “ME” means mean error.
\end{minipage}
\end{table}

In the configuration without prompts, SAM failed to identify cracks in full-scene images containing diverse background elements. The segmentation output only distinguished between major structural components and non-concrete regions, but was unable to detect cracks embedded within concrete surfaces due to their subtle and context-dependent nature, resulting in a IoU of 2\%.
As shown in Figure~\ref{fig:sam picture}b, the fractal dimension matrix-based prompt achieves acceptable segmentation performance on localized images that contain only cracks and homogeneous concrete surfaces. However, in more complex scenes-including structural edges, stains, occlusions, or background clutter-this method lacks feature selectivity. In such cases, the prompts often activate on both crack and non-crack features (e.g., surface noise or unrelated objects), resulting in segmentation failure (<5\% in IoU). These findings highlight the limitations of prompt strategies based on full-scene image pre-processing in real-world applications.
In contrast, crack-aware prompts generated via pre-trained models exhibit greater sensitivity to crack-relevant patterns. Nevertheless, without a dedicated quality assessment mechanism, these prompts may cause over-segmentation in scenarios involving parallel cracks or interfering features, resulting in a IoU of 22.9\%. The integration of a segmentation quality assessment module significantly mitigates this issue, resulting in more precise and reliable crack masks (53.2\% in IoU).

In general, it can be noticed that the proposed methods have a significant performance improvement (up to 50\% IoU compared with foundation model SAM without specially designs).
The combined design of crack-aware prompting and quality evaluation enables SAM to perform effective crack segmentation not only in localized regions, but also in full-scene images that include occlusions, stains, and surrounding structural elements. The proposed prompts are selectively responsive to crack features, effectively suppressing irrelevant background interference and enhancing segmentation robustness under complex field conditions.

\subsection{Results of Crack Localization and Automated Crack Measurement}\label{sec5.2}
This section presents the field experiment results for 3D reconstruction, crack localization, and automated width measurement.
As described in Section~\ref{sec6}, the reconstruction process begins with large-scale scene mapping using LiDAR-SLAM, followed by point cloud filtering and smoothing using SOR and MLS. Figure~\ref{fig:reconstruction}a illustrates the full reconstruction pipeline. For targeted evaluation, the reconstructed point cloud was cropped to include only the concrete specimens.

Subsequently, multi-sensor fusion was performed following the procedure detailed in Section~\ref{sec7.1}. Keyframes were automatically selected, and corresponding images were projected onto the point cloud to generate a 3D colorized structure model. 
Crack masks were also projected in the same manner to segment crack regions, resulting in a 3D semantic crack model. The final result, shown in Figure~\ref{fig:reconstruction}b, is a combined 3D semantic and colorized crack model that enables precise spatial localization of crack information.

With the 3D semantic crack model established, the widths of 15 annotated cracks were measured using three approaches: (1) microscope-based manual measurements as reference values, (2) manual measurement on the point cloud, and (3) automated measurement following the workflow described in Section~\ref{sec7.2}.
A visual comparison of the results is presented in Figure~\ref{fig:reconstruction}c.

Table~\ref{tab:width} summarizes the automated measurement results against the reference values.
Overall, the proposed framework achieved high measurement accuracy, with an average absolute error below 0.1~mm and an average relative error under 15\%.
The maximum relative error (33.92\%) was observed for Crack~\#9, whose reference width was only 0.45~mm, corresponding to an absolute error of 0.15~mm.
Such large relative errors are mainly associated with very small reference widths, where even minor absolute deviations can lead to inflated percentage errors.
For cracks narrower than 0.5~mm (e.g., \#3, \#8, \#9, \#10, and \#13), relative errors are therefore higher due to the small denominators, while all absolute errors remain below 0.2~mm.
These results confirm the feasibility and effectiveness of the proposed method for automated crack localization and geometric measurement in practical engineering scenarios.

From a methodological perspective, residual measurement errors arise from accumulated uncertainties in pixel-level segmentation, SLAM-based localization, projection alignment, and algorithmic inference.
In addition, the achievable measurement precision is inherently constrained by image spatial resolution at the crack location, which is distance-dependent.
With the 4096$\times$3000 imaging setup used in this study, the horizontal sampling resolution is approximately 0.069~mm/pixel at an acquisition distance of 0.3~m and decreases to approximately 0.113~mm/pixel at 0.5~m.
This indicates that long-range image acquisition will reduce the attainable sub-millimeter precision unless higher optical resolution is employed.
Meanwhile, adverse imaging conditions such as motion blur or insufficient illumination in real environments may affect the accuracy of crack segmentation. This can lead to inaccurate localization of crack boundaries in images, which may subsequently propagate into width estimation.

\begin{table}[t]
\centering
\caption{Comparison of 3D reconstruction performance across SfM, LOAM, and the proposed approach.}
\label{tab:reconstruction_compare}

\footnotesize
\setlength{\tabcolsep}{4pt}
\renewcommand{\arraystretch}{1.10}

\begin{threeparttable}
\begin{tabular}{@{}%
>{\raggedright\arraybackslash}p{0.30\columnwidth}%
>{\centering\arraybackslash}p{0.20\columnwidth}%
>{\centering\arraybackslash}p{0.20\columnwidth}%
>{\centering\arraybackslash}p{0.20\columnwidth}@{}}
\toprule
\textbf{Evaluation Metric} & \textbf{SfM} & \textbf{LOAM} & \textbf{Ours} \\
\midrule

\multicolumn{4}{@{}l}{\textbf{Point Cloud Quality}} \\
\makecell[l]{Point surface density\tnote{1}\,$\uparrow$\\(pts/m$^2$)}
& 165,208.47 & 1,471,886.13 & 3,173,144.75 \\
\makecell[l]{Surface roughness\,$\downarrow$\\(mm)}
& 0.279 & 1.556 & 0.494 \\
\addlinespace[2pt]

\multicolumn{4}{@{}l}{\textbf{Geometry Accuracy}} \\
\makecell[l]{Geometric dimension\\(mm)}
& N/A & 393.9 & 380.5 \\
\makecell[l]{Mean error\,$\downarrow$\\(\%)}
& N/A & 3.66 & 0.13 \\
\addlinespace[2pt]

\multicolumn{4}{@{}l}{\textbf{Efficiency}} \\
\makecell[l]{Computational time\,$\downarrow$\\(min)}
& 300.32 & 11.2 & 35.13 \\
\addlinespace[2pt]

\multicolumn{4}{@{}l}{\textbf{On-site Inspection}} \\
Sensor types & Camera & LiDAR \& IMU & Camera \& LiDAR \& IMU \\
Real-time reconstruction & No & Yes & Yes \\
\bottomrule
\end{tabular}

\begin{tablenotes}[flushleft]
\footnotesize
\item[1] Surface density is computed with a neighborhood radius of 10~mm.
\end{tablenotes}
\end{threeparttable}
\label{tab:reconstruction compare}
\end{table}

\begin{figure*}[!tbp]
    \centering
    \includegraphics[width=0.9\linewidth]{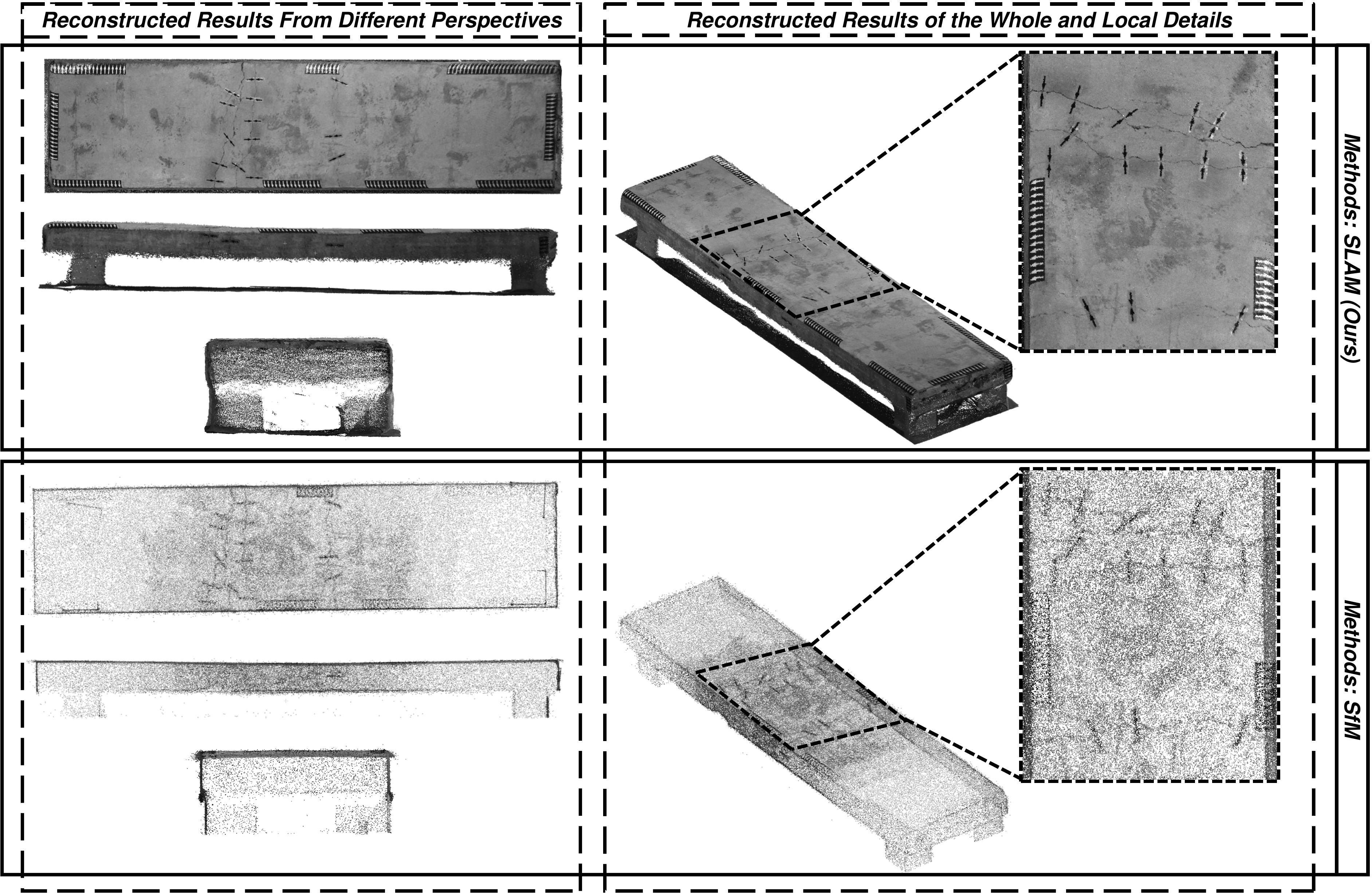}
    \caption{Comparison of 3D reconstruction results between our method and the SfM-based COMLAP method.}
    \label{fig:3D reconstruction comparison}
\end{figure*}

\subsection{Comparative Analysis of 3D Reconstruction Methods}

To evaluate the performance of the proposed multi-sensor fusion framework in terms of geometric accuracy and robustness, a comparative analysis was conducted using the same dataset across three representative 3D reconstruction methods: (1) the SfM pipeline implemented via COLMAP \citep{schoenberger2016sfm,schoenberger2016mvs,zeng_systematic_2024}, (2) the LiDAR Odometry and Mapping (LOAM) method \citep{Zhang2014LOAMLO}, and (3) the proposed method integrating LiDAR, camera, and IMU data. The comparison, detailed in Table~\ref{tab:reconstruction compare}, focuses on four core metrics: geometric accuracy, point cloud density, surface roughness, and on-site operational efficiency.
It should be emphasized that these methods rely on different sensing modalities by design.
The goal of this comparison is not to claim superiority due to the use of additional sensors, but to analyze how different sensing paradigms affect reconstruction fidelity and practicality under comparable data acquisition conditions.

In terms of geometric accuracy, precise reconstruction is essential for effective crack localization and measurement. A pedestrian walkway slab width was measured from each reconstructed point cloud and compared with manual ground-truth measurements. As indicated in Table~\ref{tab:reconstruction compare}, the SfM approach lacks absolute scale calibration, rendering geometric comparisons infeasible. In contrast, the proposed method demonstrates superior fidelity, achieving sub-millimeter accuracy and outperforming the LOAM-based method.

Point cloud density influences the detectability of fine crack details.
SfM reconstructions depend heavily on image resolution, while LOAM lacks semantic crack information due to its LiDAR-only setup. The proposed method integrates high-resolution image and LiDAR data, resulting in a significantly denser and semantically rich point cloud. The visual improvements in density are illustrated in Figure~\ref{fig:3D reconstruction comparison}.
This improvement does not stem from using more sensors per se, but from leveraging complementary sensing characteristics: LiDAR provides accurate metric geometry, while images contribute high-resolution semantic details.

Surface roughness was assessed by computing the deviation of each point from the best-fit plane derived from its local neighborhood. This metric reflects local geometric consistency and directly affects crack width estimation reliability. The results in Table~\ref{tab:reconstruction compare} indicate that the proposed method achieves smoothness on par with SfM, while significantly outperforming LOAM.

Regarding on-site efficiency, the proposed method supports real-time sparse reconstruction through tightly integrated LiDAR and IMU sensors. This capability significantly reduces the risk of reconstruction failure and eliminates the need for re-acquisition in field environments. In contrast, SfM methods require dense image overlap and controlled data capture, making them sensitive to acquisition conditions and limiting their robustness in complex engineering scenarios.

Overall, this comparison highlights the practical trade-offs among image-only, LiDAR-only, and multi-modal crack reconstruction paradigms in the context of crack inspection.
The results demonstrate that LiDAR-image fusion provides a favorable balance between geometric accuracy, semantic richness, and field robustness for millimeter-level crack measurement tasks.

\section{Conclusions}\label{sec10}

This study presents a field-deployable framework for high-precision crack inspection in concrete structures by integrating robust crack segmentation, multi-modal sensing, and 3D geometric analysis. Implemented on a portable platform combining a camera, LiDAR, and IMU, the proposed system enables accurate and automated crack inspection under complex real-world conditions. The main contributions are summarized as follows:

\begin{enumerate}
\item A crack-aware prompt generation strategy with a segmentation quality assessment module is proposed to effectively extend foundation models to crack segmentation, improving robustness and generalization in complex and unseen scenarios. Extensive experiments on real scene demonstrate a 6\% improvement in segmentation IoU under zero-shot conditions.
\item A crack-oriented, multi-frame multi-modal fusion framework is proposed for 3D crack reconstruction. By integrating RGB imagery and LiDAR point clouds, the framework generates dense and semantically enriched point clouds at real-world scale, capturing crack geometry and spatial distribution with high fidelity.
\item An automated 3D measurement pipeline is introduced to directly quantify crack width and spatial location coordinates from the reconstructed semantic point cloud, achieving a mean crack width estimation error below 0.1~mm.
\end{enumerate}

Overall, the proposed framework represents a meaningful step toward scalable, high-precision, and field-ready crack inspection for concrete infrastructure. Despite the encouraging results, several directions remain for future research.

First, while the current system is implemented on a handheld platform to verify feasibility, future work will explore deployment on UAVs to improve accessibility and efficiency for large-scale structures such as bridges and tunnels.

Second, to address challenging imaging conditions commonly encountered in practice such as low illumination and camera shake, future studies will investigate the integration of hardware-level stabilization, adaptive exposure control, and enhanced segmentation robustness.

Third, the achievable accuracy of crack width measurement is inherently constrained by the acquisition distance. For large-scale structures, exhaustive close-range scanning of the entire structure is inefficient. Future work will therefore explore adaptive scanning and path-planning strategies that combine coarse global reconstruction with locally refined data acquisition, enabling efficient multi-scale structural modeling.

Finally, while crack width and spatial location can be accurately measured within the current framework, crack length is not explicitly quantified. Since the engineering significance of crack length mainly lies in its temporal evolution, future research will focus on long-term crack tracking strategies to characterize crack propagation behavior over time.


\section*{Acknowledgment}
This work was supported by National Natural Science Foundation of China (project No. 52308225, 2022HWYQ04) and Natural Science Foundation of Hunan Province, China (project No. 2023JJ40721). Their support is gratefully acknowledged. The authors also acknowledge Prof. Takashi Matsumoto from Hokkaido University for fruitful discussions.
\bibliography{citation}
\end{document}